\documentclass[journal]{IEEEtran}
\ifCLASSINFOpdf
\else
\fi
\usepackage{orcidlink}
\usepackage{cite}
\usepackage{cuted}
\usepackage{array}
\usepackage{amsmath,amssymb,amsfonts}
\usepackage{amsthm}
\usepackage{makecell}
\usepackage{algorithmic}
\usepackage{graphicx}
\usepackage{subcaption}
\usepackage{amsmath}
\usepackage{textcomp}
\usepackage{xcolor}
\usepackage{float}
\usepackage{stfloats}
\newtheorem{theorem}{\bf {Theorem}}
\def\BibTeX{{\rm B\kern-.05em{\sc i\kern-.025em b}\kern-.08em
    T\kern-.1667em\lower.7ex\hbox{E}\kern-.125emX}}

\begin{document}
%
\title{Reduced Effectiveness of Kolmogorov-Arnold Networks on Functions with  Noise}
%
%
%

\author{Haoran~Shen,
        Chen~Zeng,
        Jiahui~Wang,
    and~Qiao~Wang\orcidlink{0000-0002-5271-0472},~\IEEEmembership{Senior Member,~IEEE}%
 
\thanks{Both H. Shen and C. Zeng was with the School of Information Science and Engineering, Southeast University, Nanjing, China (email: haoranshen28@163.com, zengchen0797@163.com). }
\thanks{J. Wang was with the School of Economics and Management, Southeast University, Nanjing, China (email: wangjh0512@qq.com).}
\thanks {Q. Wang was with both the School of Information Science and Engineering and the School of Economics and Management, Southeast University, Nanjing, China (Corresponding Author, email: qiaowang@seu.edu.cn).}  
}

\maketitle

\begin{abstract}
It has been observed that even a small amount of noise introduced into the dataset can significantly degrade the performance of KAN. In this brief note, we aim to quantitatively evaluate the performance when noise is added to the dataset. We propose an oversampling technique combined with denoising to alleviate the impact of noise. Specifically, we employ kernel filtering based on diffusion maps for pre-filtering the noisy data for training KAN network. Our experiments show that while adding i.i.d. noise with any fixed SNR, when we increase the amount of training data by a factor of $r$, the test-loss (RMSE) of KANs will exhibit a performance trend like $\text{test-loss} \sim \mathcal{O}(r^{-\frac{1}{2}})$ as $r\to +\infty$. We conclude that applying both oversampling and filtering strategies can reduce the detrimental effects of noise. Nevertheless, determining the optimal variance for the kernel filtering process is challenging, and enhancing the volume of training data substantially increases the associated costs, because the training dataset needs to be expanded multiple times in comparison to the initial clean data. As a result, the noise present in the data ultimately diminishes the effectiveness of Kolmogorov-Arnold networks.
\end{abstract}

\section{Introduction}
The Kolmogorov-Arnold networks (KAN) have attracted considerable attention following their release on Arxiv \cite{liu2024kan}. However, \cite{Zhang2024} pointed out that these networks are susceptible to noise.

Having been introduced only a few months ago, KANs are considered innovative neural network structures and potential substitutes for Multi-Layer Perceptrons (MLPs). Various applications of KAN networks have been reported across different domains, including: time series analysis\cite{2024arXiv240508790V}\cite{2024arXiv240602496X}, ODEs\cite{2024arXiv240704192K}, PDEs\cite{2024arXiv240611045W}, hyperspectral image classification\cite{2024arXiv240607869T}\cite{2024arXiv240615719J}, physical modeling\cite{2024arXiv240507488P}, computer vision\cite{2024arXiv240609087A}\cite{2024arXiv240602918L}, and graph learning\cite{2024arXiv240618380B}\cite{2024arXiv240613597Z}\cite{2024arXiv240606470K}.

In addition, various enhancements to KANs have been introduced. For instance, \cite{2024arXiv240507200S} substituted the spline functions used as weights with Chebyshev polynomials, \cite{2024arXiv240507344G} merged the strengths of KANs and LSTM, \cite{2024arXiv240512832B} utilized wavelet-based structure for KANs, \cite{2024arXiv240613155D} developed Convolutional KANs, \cite{2024arXiv240607456A} employed fractional-orthogonal Jacobi functions as the basis functions for KANs, and \cite{2024arXiv240602075Q} enhanced the computational process of KANs.

In this paper, we aim to measure the notable decrease in performance of KAN networks when subjected to noise interference, and explore methods to alleviate the noise effects. We propose two strategies: implementing filtering techniques and increasing the volume of training data.

The contributions of this paper are as follows: Firstly, it demonstrates that incorporating noise into the training data drastically diminishes the performance of Kolmogorov-Arnold Networks (KANs). To tackle this problem, the authors propose two strategies. One strategy utilizes a kernel filtering technique to mitigate some of the noise. The difficulty lies in determining the optimal variance parameter for the filter since it's nonlinearly dependent on the Signal-to-Noise Ratios (SNRs). The other strategy addresses the noise issue by expanding the training dataset. The authors discovered an intriguing pattern showing that if the number of training samples is increased by a factor of $r$ from the initial amount, the performance (test-loss) will asymptotically decrease as $r^{-\frac{1}{2}}$ as $r \to \infty$.

The structure of this paper is organized as follows: In Section \ref{S-A}, we showed that introducing noise to the training dataset leads to a significant drop in the performance of KANs. We then develop two methods to counteract the noise effect. In Section \ref{S-B}, we propose a filtering technique based on kernel filtering, which can remove some noise to an extent. The most difficult challenge here is determining the optimal variance parameter $\sigma$ of this Gaussian-like kernel filter. Unfortunately, the optimal variance is nonlinearly dependent on the SNRs, making it difficult to ascertain. Following this, in Section \ref{S-C}, we introduce a technique to reduce noise interference by increasing the training dataset size and demonstrate how the test loss statistically declines asymptotically as the data volume increases.  In Section \ref{SCM}, we integrate kernel filtering with oversampling. However, it turns out to be challenging to find an equilibrium between the repetition factor $r$ of the data size and the SNR of the data. Finally, in Section \ref{S-E}, we draw our conclusions for this paper.

\section{The impact of noise in KANs}\label{S-A}
The Kolmogorov-Arnold theorem addresses the representation of multivariate continuous functions. The theorem states that any continuous function of multiple variables can be represented as a superposition of continuous functions of one variable and addition \cite{Kolmogorov1956-2} \cite{Kolmogorov1957} \cite{Arnold1957}. Formally, it can be stated as:

\begin{theorem}[Kolmogorov-Arnold Theorem]
Let $f:\ [0,1]^n\to \mathbb R$ be any multivariate continuous function, there exist continuous univariate functions \( \phi_i \) and \( \psi_{ij} \) such that:

\begin{equation}\label{ka-1}
f(x_1, x_2, \ldots, x_n) = \sum_{i=1}^{2n+1} \phi_i \left( \sum_{j=1}^{n} \psi_{ij}(x_j) \right).
\end{equation}
\end{theorem}

Unfortunately, this theorem is existential rather than constructive, unlike the Lagrange interpolation theorem. In 2009, \cite{Braun2009} provided a constructive proof of this theorem. Nonetheless, it might pose challenges when working with functions that exhibit noise.

According to \cite{Zhang2024}, adding a small Gaussian noise to the training labels of a dataset and employing this altered dataset to train a KAN network can unexpectedly worsen the test loss.

To ascertain the existence of this phenomenon and explore solutions to reduce noise impact, we added noise to the training  datasets involved in different fitting tasks. These noisy datasets were subsequently fed into the network for training, and the results were contrasted against those obtained from the original datasets. We selected six functions for fitting and utilized various KAN network architectures. They are listed in Table \ref{tab_func}.
{
\begin{table*}[htbp]
\renewcommand\arraystretch{1.7}
\caption{Functions Used for Fitting and Corresponding KAN Structures}
\label{tab_func}
\begin{center}
\begin{tabular}{|c|c|c|c|}
\hline
  \textbf{Functions}   & \textbf{Structure } \\
  \hline
 $ f_1  = \exp ( {\sin \left( \pi x \right) +y^2}) $    &[2,5,1]\  cf. \cite{liu2024kan}\\
 \hline
 $ f_2 =  xy $ &  [2,5,1] \\
 \hline 
 $ f_3  = \exp ( \frac{1}{2}(\sin \left( \pi x_1^2 + \pi x_2^2 \right) + \sin \left( \pi x_3^2 + \pi x_4^2 \right) )) $ & [4,2,1,1]\\ \hline
  $ f_4 = 1 + x\sin(y) $ &  [2,2,1]\\
  \hline
  $ f_5  =  \arcsin(x\sin(y)) $ &  [2,2,1]\\
  \hline
 $ f_6  = x\sqrt{y^2 + z^2} $ & [3,2,2,1]\\ 
\hline
\end{tabular}
\end{center}
\end{table*}

}

As shown in Table \ref{tab1}, it is clear that even small amounts of noise can significantly impact network performance. With 3000 training samples, adding noise with a standard deviation of $\sigma=0.2$ leads to a significant increase in test-loss.

\begin{table*}[htbp]
\renewcommand\arraystretch{1.7}
\caption{Test loss comparison between data affected by Gaussian noise ($\mu=0$, $\sigma=0.2$) and noise-free data.}
\begin{center}
\begin{tabular}{|c|c|c|c|}
\hline

\textbf{Functions} & \textbf{ Noisy-free Test Loss}& \textbf{Noisy  Test Loss}
&\textbf{SNR (dB)}\\
\hline
$f_1$ &
$5.46\times10^{-3}$&$3.20\times10^{-2}$&21.5 \\
\hline
$f_2$ &
$1.31\times10^{-3}$&$3.16\times10^{-2}$&4.7  \\
\hline
$f_3$ &
$4.22\times10^{-3}$&$3.39\times10^{-2}$&18.7\\
\hline
\end{tabular}
\label{tab1}
\end{center}
\end{table*}
What causes KAN's vulnerability to noise? The lack of regularity induced by noise leads to a deterioration in the performance of KANs. The activation function in KAN comprises a basis function $b(x)$ and a spline function \text{spline}(x), as shown in  following equation, 
\begin{equation} \label{neuron-1}
\phi \left( x \right) =w \left( b(x) +\text{spline} \left( x \right) \right),
\end{equation}
where
\begin{equation}
    b(x)=\frac{x}{1+\exp(-x)},
\end{equation}
both of which possess adequate smoothness. Moreover, KAN neurons perform only simple summation operations, making it hard to accurately represent a non-smooth function with added noise using nested smooth functions. This results in a poor recovery of the original function in a noisy environment.

\section{Denoise by Kernel filtering}\label{S-B}
We now aim to reduce the impact of noise in general multivariate functions. Given that the training data is not uniformly sampled in this scenario when fitting multidimensional functions with KANs, it is logical to utilize multidimensional filtering techniques. 

\subsection{Kernel Filtering}
In this subsection, we utilize kernel filtering to remove noise from the data. Alternatively, for data on non-linear manifolds, diffusion maps can be employed as described in \cite{coifman2006diffusion}.

For the sake of simplicity, we 
consider the Gaussian-like kernel\cite{kernel2024}
\begin{equation}\label{kernel}
k(x,y)=Ce^{-d^2 \left( x,y \right) / 2 \sigma ^2} ,\end{equation}
where $C$ is the normalization coefficient, and $d$  the distance function defined as
\begin{equation}
d^2(x,y)= \sum_{n=1}^{N} { \left( x^{(n)} - y^{(n)} \right)}^2,\end{equation}
for data $ x=(x^{(n)})_{n=1}^{N},\  y=(y^{(n)})_{n=1}^N $. 
 Clearly, the kernel filtering for data $\{f(x_n);\ n=1,2,\cdots,M\}$ produces up-dated data
\begin{equation}
    \widehat f(x_j) = \sum_{n=1}^M  k (x_j,x_n) f(x_n),
\end{equation}
and the kernel $k(x_j,x_n)$ might be stored as a matrix in practical computation. Actually, for large matrix, we may treat it as sparse matrix since $k(x_j,x_n)\approx 0$ when $d(x_j,x_n)$ is large enough.

Notice that the standard deviation-like $\sigma$ is  the scaling parameter, and might be regarded as the "width" of kernel function.

\begin{figure*}[htbp]
\centering
  \subfloat[\centering
  $f_2$
  ]
  {
    \includegraphics[width=0.49\linewidth]{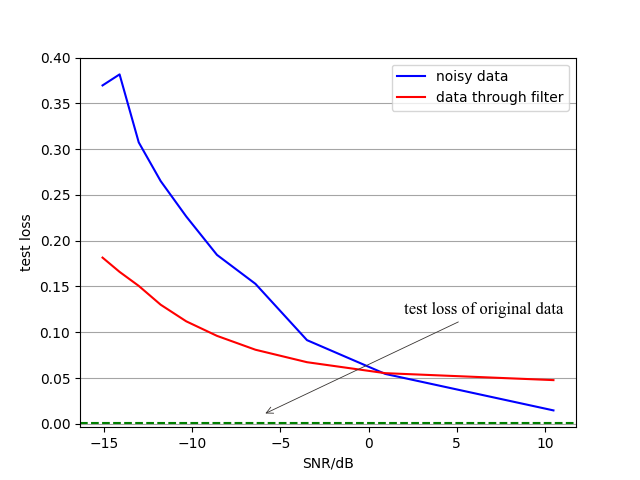}
  }
  \subfloat[\centering
  $f_3$
]
  {
    \includegraphics[width=0.49\linewidth]{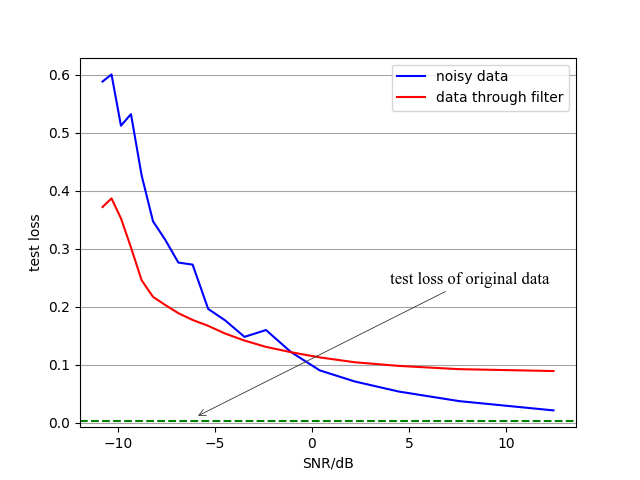}
  }
  \caption{Applying kernel filtering to $f_2$ and $f_3$ with $\sigma=0.1$.
  }
  \label{fig1}
\end{figure*}
We will evaluate the performance of the previously mentioned functions $f_2$ and $f_3$ by altering the SNRs using kernel filtering with $\sigma =0.11$. To obtain the data presented in Figure \ref{fig1}, we conducted this experiment three times and calculated the average of the results.

For $f_2$, as the SNR increases, both noisy and filtered data will exhibit a reduced test loss, with an intersection occurring at an SNR of $0$dB. This indicates that kernel filtering can be beneficial for contaminated datasets when the SNR is below $0$dB, but it becomes detrimental when the SNR exceeds $0$dB. In the case of $f_3$, the effect is comparable, except the intersection point shifts to $-2$dB. This suggests that in low SNR scenarios, kernel filtering can slightly enhance the performance of KAN networks. 

\begin{figure}[htbp]
\centerline{\includegraphics[width=1.03\linewidth]{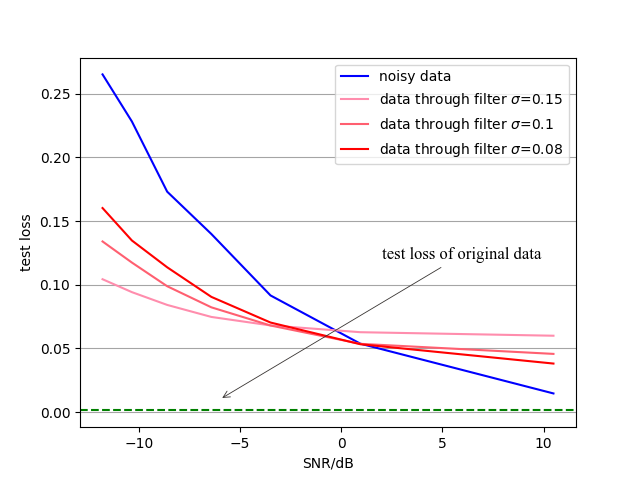}}
\caption{Applying kernel filtering to $f_2$ with different $\sigma$.}
\label{fig2a}
\end{figure}

To investigate the impact of $\sigma$ on filtering performance, we employed $f_2$ as the fitting objective and injected Gaussian noise denoted by $N(0,0.2)$. Various levels of kernel filtering were implemented, leading to the test-loss versus SNR curves illustrated in Fig \ref{fig2a}. It is evident that beginning with $\sigma=0.08$ (small $\sigma$), the curve resembles noisy data shown in blue and diminishes rapidly. As $\sigma$ increased (with a larger $\sigma$ resulting in smoother data), the initial point on the left side of the performance curve improves, achieving lower test-loss initially. Nevertheless, the decay is more gradual, and importantly, the right side of the curve performs worse, displaying a considerably higher test loss compared to earlier curves. {\bf Ultimately, no matter what $\sigma$ value is selected, kernel filtering ceases to function effectively once the data's SNR surpasses the critical limit. Filtering is only effective in the low SNR region.} 

%


\begin{figure*}[!htbp]
\centering

  \subfloat[\centering
  Fitting $f_4$ with training samples = 500
  \label{f4_r1}]
  {
    \includegraphics[width=0.33\linewidth]{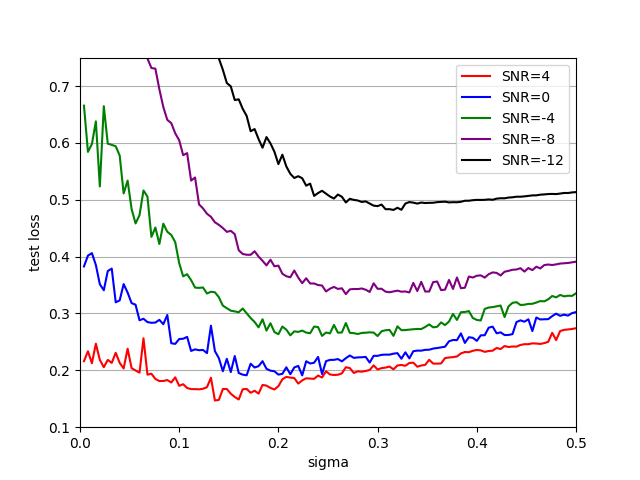}
  }
  \subfloat[\centering
  Fitting $f_5$ with training samples = 500
  \label{f5_r1}]
  {
    \includegraphics[width=0.33\linewidth]{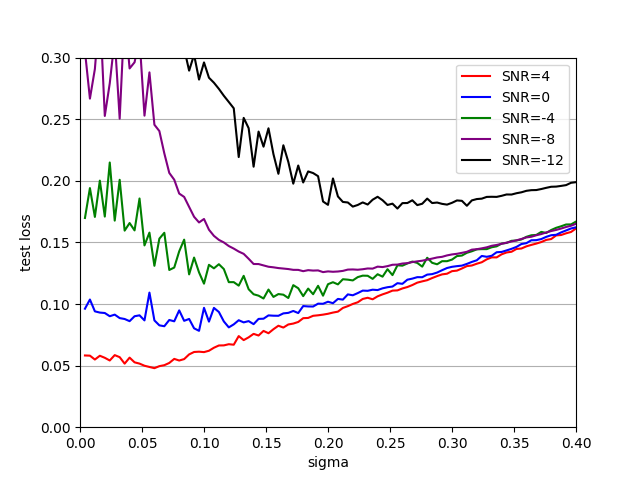}
  }
  \subfloat[\centering
  Fitting $f_6$ with training samples = 500
  \label{f6_r1}]
  {
    \includegraphics[width=0.33\linewidth]{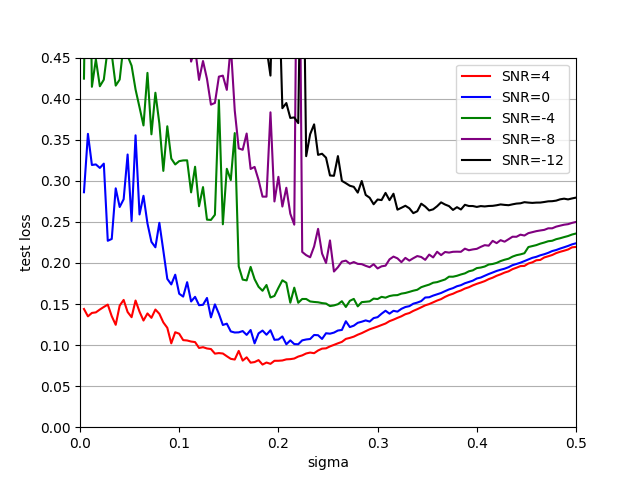}
  }

  \subfloat[\centering
  Fitting $f_4$ with training samples = 5000
  \label{f4_r10}]
  {
    \includegraphics[width=0.33\linewidth]{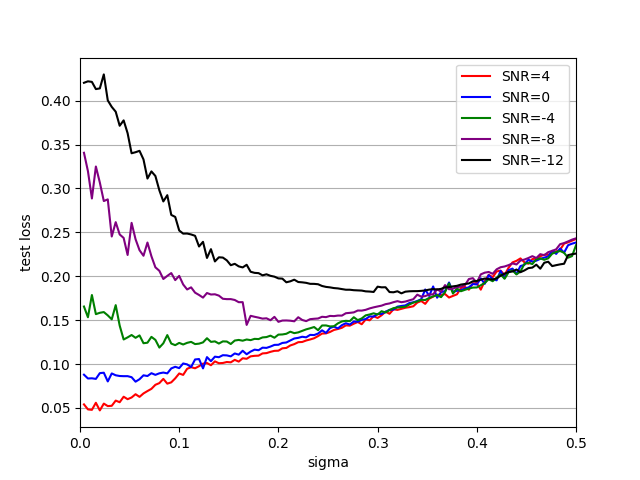}
  }
  \subfloat[\centering
  Fitting $f_5$ with training samples = 5000
  \label{f5_r10}]
  {
    \includegraphics[width=0.33\linewidth]{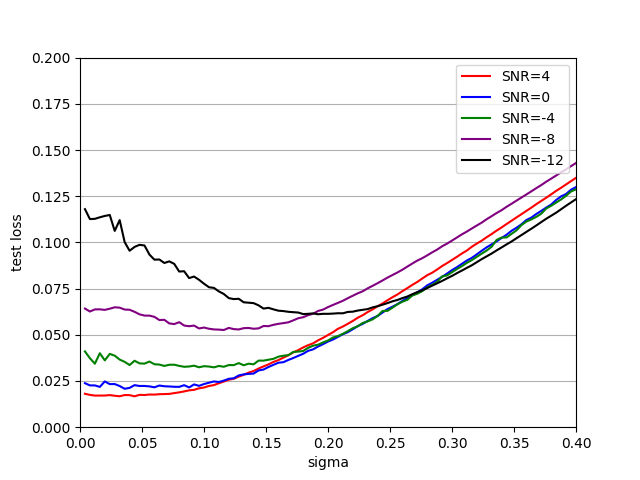}
  }
  \subfloat[\centering
  Fitting $f_6$ with training samples = 5000
  \label{f6_r10}]
  {
    \includegraphics[width=0.33\linewidth]{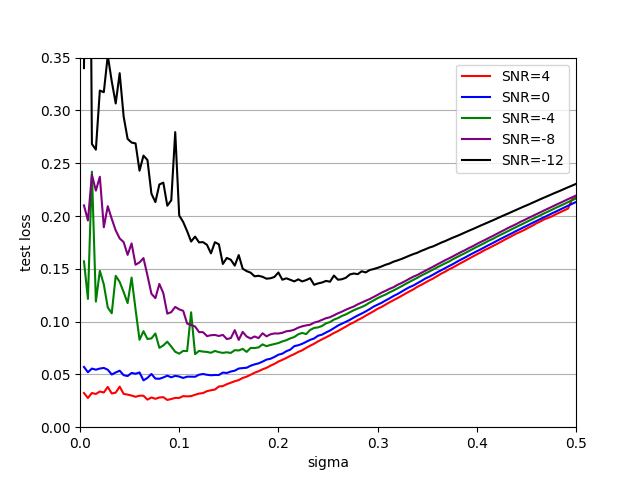}
  }
  \caption{Filtering performance with different values of $\sigma$ under various SNRs.
  }
  \label{f456}
\end{figure*}

\subsection{The Optimal Filter Parameter $\sigma$}
Let's focus on the equation \eqref{kernel} that describes Gaussian-like kernel. The parameter $\sigma$ represents the standard deviation of the Gaussian function, which controls the width of the kernel. 

To find the impact of $\sigma$ on filtering performance, We chose functions $f_4,f_5$ and $f_6$ for fitting. We applied Gaussian-like kernel filtering with different values of $\sigma$ under various SNR conditions for the three functions mentioned above. The number of training samples is $500$. The results depicted in Fig \ref{f4_r1}, \ref{f5_r1} and \ref{f6_r1}, show that as $\sigma$ increased, the test loss initially decreased and then increased, indicating the existence of a best $\sigma$. Additionally, with the decrease in SNR, the best $\sigma$ tends to increase.

The results are straightforward to understand. When $\sigma$ is very small, the filtering isn’t sufficient and does not smooth the noise effectively. On the other hand, if $\sigma$ is too large, excessive filtering happens, which can obscure critical details in the data. Additionally, as the SNR diminishes, indicating higher noise levels, the standard deviation of Gaussian noise increases. Therefore, the optimal $\sigma$ for the Gaussian-like kernel filter must also increase to efficiently mitigate the higher noise levels. 
However, the best value for $\sigma$ differed among various functions ($f_4$, $f_5$, and $f_6$) and SNRs. Additionally, in real-world situations, the SNR is usually not known, complicating the task of finding the optimal $\sigma$ to attain the best filtering results for general functions with noise of unknown variance.

\section{Enhance training dataset to mitigate noise}\label{S-C}
In contrast to denoising techniques, we discovered that augmenting the number of training samples is an efficient method for reconstructing noisy data. This process is akin to recovering a band-limited signal from noisy discrete oversampling data, which can be elucidated by frame theory \cite{Daubechies}.

\subsection{Reconstruct Signal from Noisy Oversampled Data} 
 Numerous important studies in signal processing have been conducted on this subject over the past decades, e.g. \cite{285644}. We outline the process using the formula
\begin{equation}\label{sampling-formula}
    f(t)=\sum_k f_k\ \text{sinc} (t-\frac{kT}{2\Omega}),\ \ f_k=f(\frac{kT}{2\Omega})+\epsilon_k
\end{equation}
where $f_k$ denotes the samples and $\epsilon_k$ signifies i.i.d. Gaussian noise with zero mean. We highlight that $0<T\le 1$ to ensure the oversampling rate, and a smaller $T$ implies a higher number of samples. Despite the presence of noise in these samples, the sampling formula \eqref{sampling-formula} acts as a stationary oversampling reconstruction method that closely approximates the original noise-free signal $f(t)$, as explained by frame theory in \cite{Daubechies}. Specifically, a higher sampling rate can mitigate the effect of samples with lower SNR. This formula \eqref{sampling-formula} is effective because the target function $f(t)$ is band-limited, implying it must be highly regular, being an entire function of exponential $|\Omega|$ type according to Paley-Wiener's Theorem \cite{Papoulis1977SignalA}.

\subsection{Increase Training Samples }\label{AA}
In our case, a larger training dataset allows KANs to extract the original data from a substantial amount of information and resist noise interference\cite{oversampling}. We discovered that increasing the quantity of training samples is an effective method for reconstructing data affected by noise. By having a larger set of training samples, KANs are capable of extracting the original data from the extensive samples and enduring noise interference.

\begin{figure*}[htbp]
    \centering
    \begin{minipage}[b]{0.49\linewidth}
        \centering
        \includegraphics[width=\linewidth]{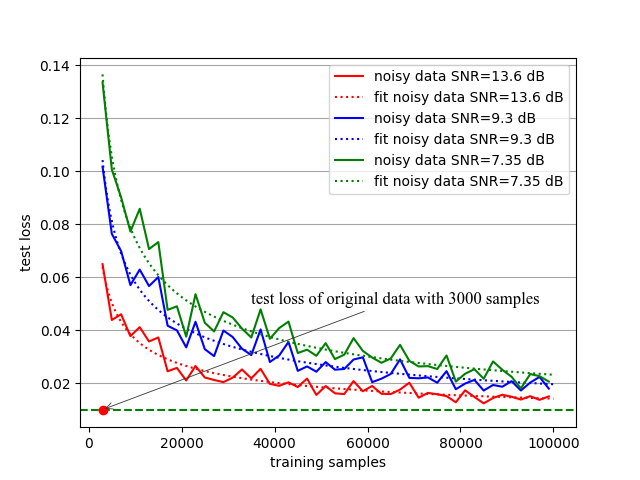}
        \caption{Applying oversamlping to $f_1$ with different SNRs.}
        \label{fig3aa}
    \end{minipage}
    \hfill
    \begin{minipage}[b]{0.49\linewidth}
        \centering
        \includegraphics[width=\linewidth]{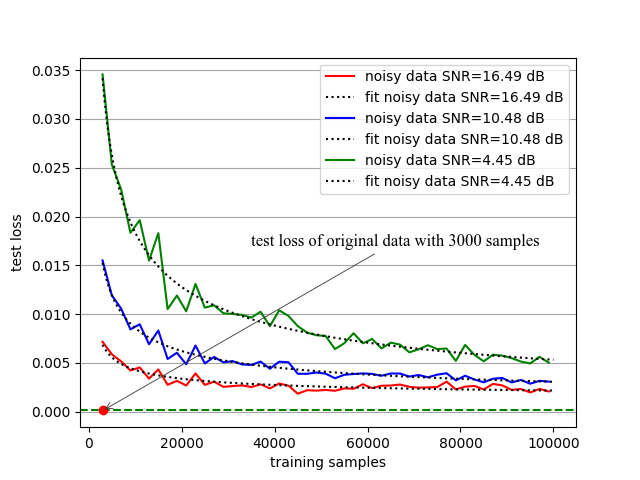}
        \caption{Applying oversamlping to $f_2$ with different SNRs.}
        \label{fig4aa}
    \end{minipage}
\end{figure*}

Using $f_1(x, y)= \exp \left( { \sin \left( \pi x \right) +y^2} \right)$ and $f_2(x, y)= xy$ as an illustration, we start with 3000 training samples and incrementally add 2000 samples at each step. These datasets are input into a [2,5,1] network at various sampling rates, and the test loss for each sample size is plotted, resulting in Figure \ref{fig3aa} and Figure \ref{fig4aa}.

\begin{figure*}[htbp]
\centering
  \subfloat[\centering
  {[4,4,2,1]} KAN
  \label{fig6aa}]
  {
    \includegraphics[width=0.49\linewidth]{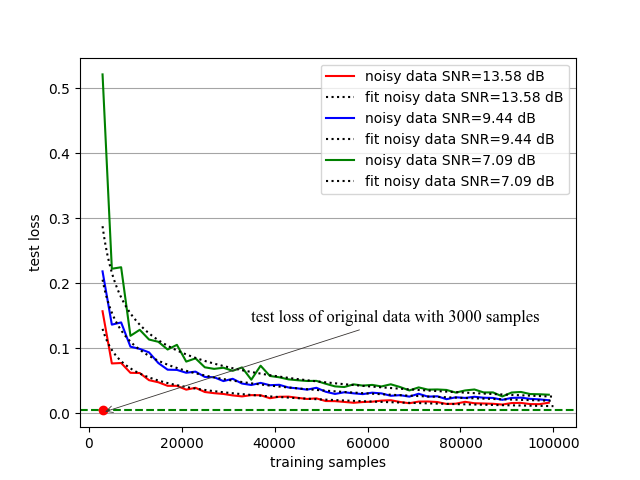}
  }
  \subfloat[\centering
  {[4,2,1,1]} KAN
  \label{fig5aa}]
  {
    \includegraphics[width=0.49\linewidth]{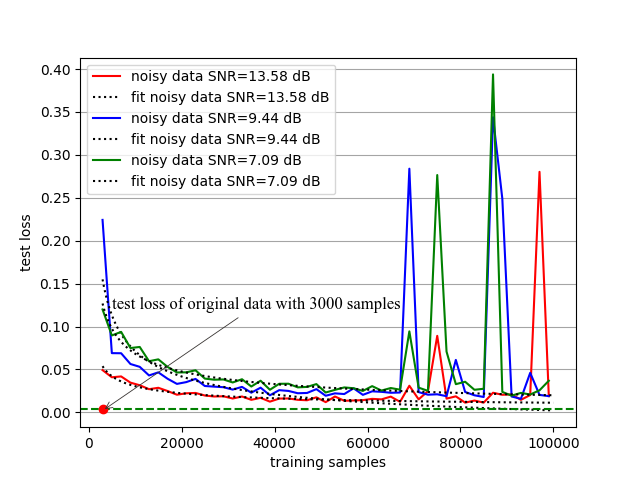}
  }
  \caption{Applying oversampling to $f_3$ with different SNRs.
  }
  \label{f3_oversampling}
\end{figure*}

To fit $ f_3 ( x_1, x_2, x_3, x_4 ) = \exp ( \frac{1}{2}(\sin \left( \pi x_1^2 + \pi x_2^2 \right) + \sin \left( \pi x_3^2 + \pi x_4^2 \right) )) $ with [4,4,2,1] KAN will result in Figure \ref{fig6aa}, which produces the same asymptotical behavior as other functions. However, if we utilize a [4,2,1,1] KAN as suggested in \cite{kernel2024}, we observe that  more data and higher SNR might exhibit higher test losses, as illustrated in Figure \ref{fig5aa}. Through extensive experimentation, we found that this pattern persisted. The [4,4,2,1] KAN shown in Fig. \ref{fig6aa} approximates $f_3$ more accurately than the [4,2,1,1] KAN illustrated in Fig. \ref{fig5aa}, reducing the considerable variations observed with the [4,2,1,1] KAN. Furthermore, as the dataset grows, the [4,4,2,1] KAN better fits with the function. It is important to note that KAN might not always adhere to the anticipated structure of the function, making pruning operations necessary for achieving a more efficient network architecture.

\begin{figure*}[htbp]
\centering
\centerline{\includegraphics[width=0.8\linewidth]{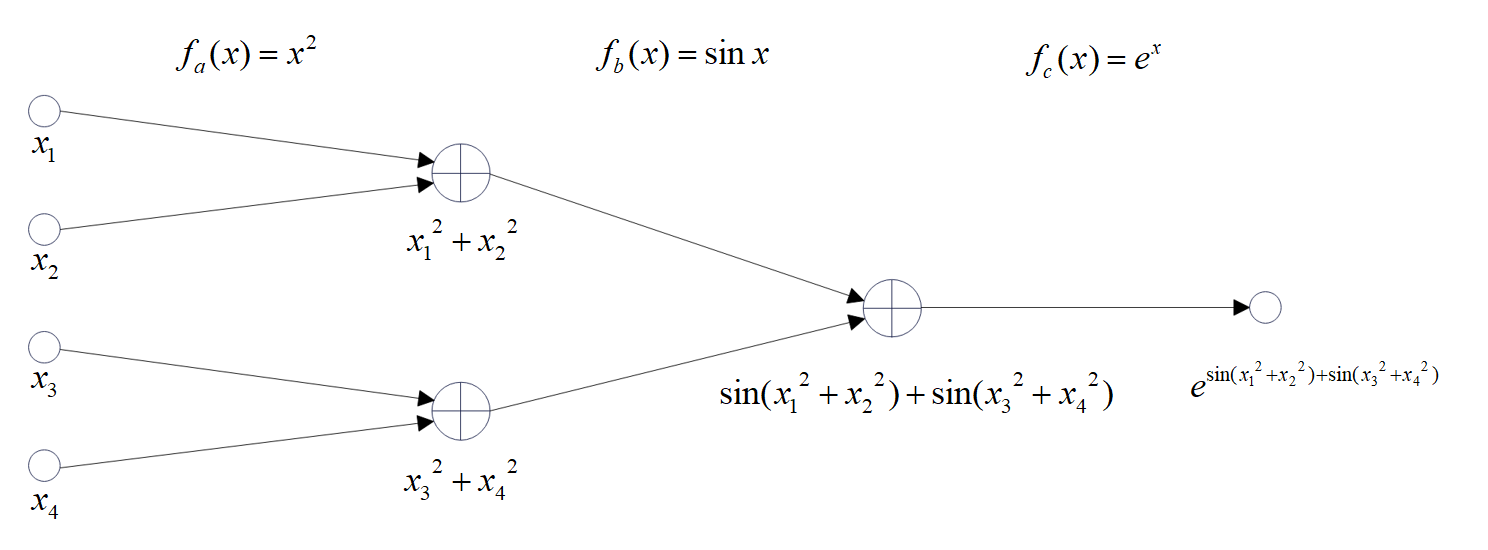}}
\caption{Ideal [4,2,1,1] KAN structure for $f_3$}
\label{f3_4211_structure}
\end{figure*}

For $f_3$, the structure [4,2,1,1] seems ideal, as shown in Fig. \ref{f3_4211_structure}. Nevertheless, the comparison between the structures [4,2,1,1] and [4,4,2,1] in Fig. \ref{f3_oversampling} reveals that KAN needs more parameters and a more complex structure to properly model this function. According to Table 3 in \cite{liu2024kan}, which compares manually designed KAN shapes with those discovered automatically, for other functions, KAN might require fewer parameters than the predetermined structure. This implies that KAN could have difficulty precisely defining the function and finding the optimal structure, hence complicating the determination of the most suitable KAN configuration for practical applications. As a result, modeling functions with KAN purely based on noisy training data without any prior understanding of the function may lead to inconsistent results.

. 


As anticipated, the test loss of KAN decreases steadily as the number of training samples grows. Initially, with the increasing sample size, the test loss drops rapidly, and then it asymptotically follows a decay rate of $r^{-\frac{1}{2}}$. Here, $r$ represents the multiple of the initial training data, and the test loss will asymptotically behave like
\begin{equation}
\text{test-loss (RMSE)} \sim \mathcal{O}(r^{-\frac{1}{2}}),\ \ \text{as}\ r \to +\infty.
\end{equation}
This behavior is actually analogous to the linear reconstruction of a band-limited signal from noisy data, see \cite{Thao-Vetterli}.

When we utilize a KAN with grid=5 to fit various functions, we aim to increase the number of training samples and generate a graph illustrating the relationship between test loss and training samples. For $f_3$, we reached a conclusion similar to $f_1$: The test loss rapidly decreases to about half under noisy conditions when the training sample size reaches 20000, and beyond that point, the rate of decline in test loss gradually slows down. The optimal denoising effect is achieved when the sample size reaches approximately 60000, where the test loss is nearly equivalent to the test loss obtained from training on the original dataset. However, while increasing the sampling rate for $f_2$ significantly reduces test loss, it remains challenging to replicate the results from training on the original dataset. Nonetheless, this outcome is still quite promising. 

\section{Combining oversampling and kernel filtering}\label{SCM}
Previous experiments have shown that both oversampling and kernel filtering are effective in reducing noise. It is natural to combine these techniques by applying kernel filtering to the dataset and slightly increasing the training sampling rate. We will then compare the test loss with that of a dataset that uses only one noise reduction method. This experiment introduces different noise levels to three functions, resulting in SNRs of $7.38$dB, $4.46$dB, and $10.53$dB for the datasets $f_1$, $f_2$, and $f_3$, respectively. The results are presented in Table \ref{tab2}

\begin{table*}[htbp]
\renewcommand\arraystretch{1.7}
\caption{Noisy test-loss  compares with noisy-free test-loss}
\begin{center}
\begin{tabular}{|c|c|c|c|c|c|c|}
\hline
\textbf{Functions} & \textbf{original dataset} & \textbf{test loss with noise}
& \textbf{kernel filtering} 
& \makecell[c]{\textbf{25 times  size of } \\ \textbf{original samples}}
& \makecell[c]{\textbf{25 times of original} \\ \textbf{kernel filtering ($\sigma=0.1$)}}
& \makecell[c]{\textbf{50 times of original} \\ \textbf{kernel filtering ($\sigma=0.1$)}}\\
\hline
$f_1$ &$4.94\times10^{-3}$ &$2.14\times10^{-1}$ &$3.39\times10^{-1}$ &$3.42\times10^{-2}$
& $2.79\times10^{-1}$ & $2.77\times10^{-2}$\\
\hline
$f_2$ &$1.31\times10^{-3}$ &$3.16\times10^{-2}$ &$5.12\times10^{-2}$
&$7.31\times10^{-3}$ &$5.52\times10^{-2}$ & $5.43\times10^{-2}$\\
\hline
$f_3$ &$4.22\times10^{-3}$ &$8.73\times10^{-2}$ &$8.69\times10^{-2}$
&$1.87\times10^{-2}$ &$1.03\times10^{-1}$ &$9.96\times10^{-2}$\\
\hline
\end{tabular}
\label{tab2}
\end{center}
\end{table*}

Regrettably, merging the two techniques did not yield significantly improved outcomes. In fact, this combined approach results in a higher test loss compared to using oversampling alone. It appears that kernel filtering disrupts the effectiveness of oversampling, and this disruption remains even when the sample size is increased to 50 times its original size.


As mentioned earlier, increasing the amount of training data can effectively reduce test loss and enhance filtering performance. Therefore, for these functions ($f_4$, $f_5$, and $f_6$), we increased the training data by 10 times, respectively, to observe how the test loss changes. The results are shown in Fig \ref{f4_r10}, \ref{f5_r10} and \ref{f6_r10}. We can find that compared to the training data size of 500, the training data size of 5000 significantly reduces the test loss. Additionally, as $\sigma$ varies, the change in test loss also decreases considerably, indicating that the impact of $\sigma$ on filtering performance is diminishing.

From the experiments performed, it is clear that both the particular function and the SNR influence the value and variability of the optimal $\sigma$, making it difficult to ascertain. In real-world situations, increasing the sampling rate is a more effective method than filtering.

\section{Conclusion}\label{S-E}
In this brief note, we assess the decline in KANs performance for functions affected by noise. We explore two methods to alleviate these problems: the first focuses on noise elimination, and the second on increasing the size of the training dataset. Although the latter method demonstrates considerable enhancement, the overall performance remains deficient due to the excessive amount of required data. Consequently, we conclude that KANs must overcome the challenges presented by noise interference.

\section*{Acknowledgment}
The authors would like to express their heartfelt gratitude to Dr.~ Aijun Zhang for the insightful discussions regarding the adverse impacts of noise on the KANs network, and for providing numerous invaluable suggestions for this manuscript.

\bibliographystyle{IEEEtran}

\bibliography{reference.bib}
\end{document}